\documentclass[runningheads]{llncs}
\usepackage{graphicx}
\usepackage{amsmath,amssymb} 
\usepackage{color}
\usepackage{float}
\usepackage[caption = false]{subfig}
\usepackage{comment}
\usepackage{amsmath,amssymb} 
\usepackage{multirow}
\usepackage{adjustbox}
\usepackage[algo2e]{algorithm2e} 
\usepackage{algorithm}
\usepackage{algorithmic}
\usepackage{placeins}
\usepackage{bbm}
\SetKwRepeat{Do}{do}{while}
\usepackage{hyperref}
\usepackage{colortbl}
\newcolumntype{P}[1]{>{\centering\arraybackslash}p{#1}}
\usepackage[font=small,labelfont=bf]{caption}
\makeatletter
\newcommand{\printfnsymbol}[1]{%
  \textsuperscript{\@fnsymbol{#1}}%
}
\makeatother
\begin{document}
\pagestyle{headings}
\mainmatter
\def\ACCV20SubNumber{975}  
\title{COMET: Context-Aware IoU-Guided Network for Small Object Tracking} 

\author{Seyed Mojtaba Marvasti-Zadeh\thanks{equal contribution}\inst{1,2,3} \and
Javad Khaghani\printfnsymbol{1}\inst{1}\and
Hossein Ghanei-Yakhdan\inst{2} \and
Shohreh Kasaei\inst{3} \and
Li Cheng\inst{1}}
\authorrunning{S. M. Marvasti-Zadeh, J. Khaghani et al.}
%
\institute{University of Alberta, Edmonton, Canada
\email{\{mojtaba.marvasti,khaghani,lcheng5\}@ualberta.ca} \and
Yazd University, Yazd, Iran \\
\email{hghaneiy@yazd.ac.ir}  \and
Sharif University of Technology, Tehran, Iran\\
\email{kasaei@sharif.edu}}
\maketitle
\begin{abstract}
We consider the problem of tracking an unknown small target from aerial videos of medium to high altitudes. 
This is a challenging problem, which is even more pronounced in unavoidable scenarios of drastic camera motion and high density. 
To address this problem, we introduce a context-aware IoU-guided tracker (COMET) that exploits a multitask two-stream network and an offline reference proposal generation strategy. The proposed network fully exploits target-related information by multi-scale feature learning and attention modules. 
The proposed strategy introduces an efficient sampling strategy to generalize the network on the target and its parts without imposing extra computational complexity during online tracking. 
These strategies contribute considerably in handling significant occlusions and viewpoint changes. 
Empirically, COMET outperforms the state-of-the-arts in a range of aerial view datasets that focusing on tracking small objects. Specifically, COMET outperforms the celebrated ATOM tracker by an average margin of $6.2\%$ (and $7\%$) in precision (and success) score on challenging benchmarks of UAVDT, VisDrone-2019, and Small-90.
\end{abstract}
\section{Introduction} \label{sec:1_Intro}
\begin{figure}[t!]
\centering
\includegraphics[width=11.5cm, height=4.9cm]{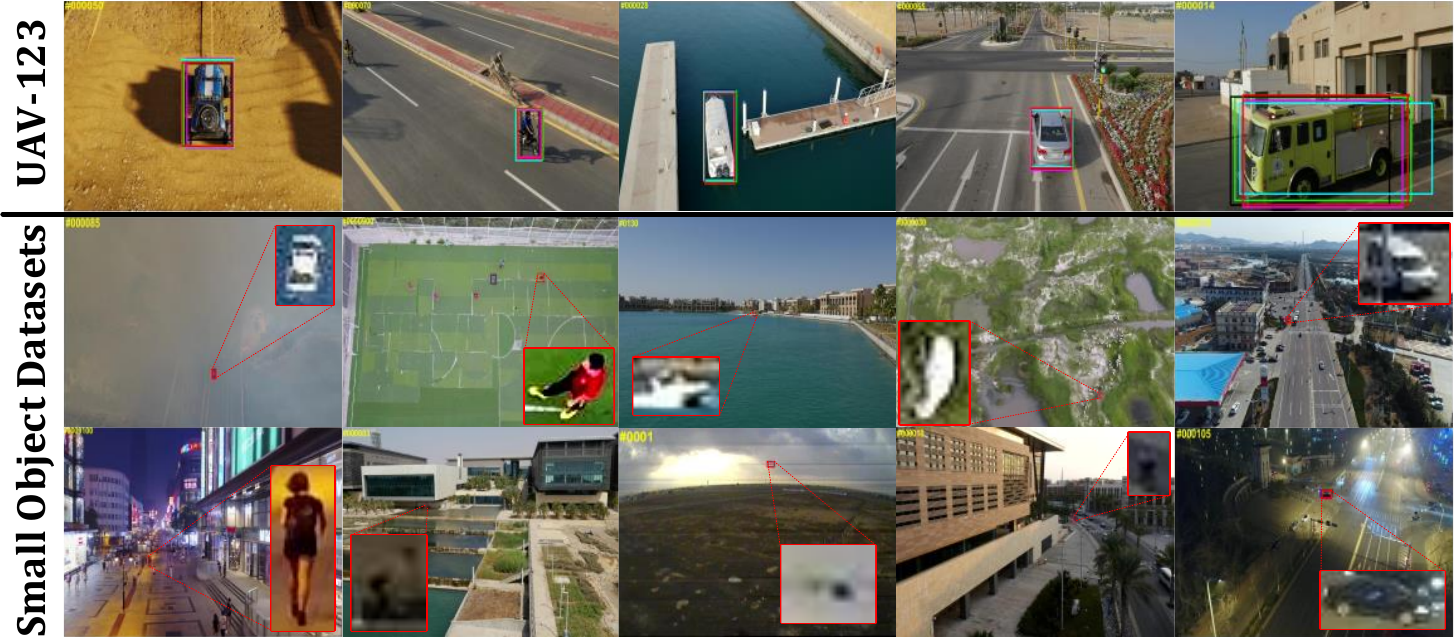}
\vspace{-2mm}
\caption{Examples to compare low-altitudes and medium/high-altitudes aerial tracking. The first row represents the size of most targets in UAV-123 \cite{UAV123} dataset, which captured from 10$\sim$30 meters. However, some examples of small object tracking scenarios in UAVDT \cite{UAVDT2018}, VisDrone-2019 \cite{VisDrone2019}, and Small-90 \cite{Small90Dataset} datasets are shown in last two rows. The UAV-123 contains mostly large/medium-sized objects, while targets in small object datasets just occupy few pixels of a frame. Note that Small-90 has been incorporated small object videos of different datasets such as UAV-123, OTB, and TC-128 \cite{TC128}. The focus on this work will be on small/tiny object tracking.}
\label{fig:Scenarios}
\vspace{-.4cm}
\end{figure}

Aerial object tracking in real-world scenarios \cite{VisDrone2019,UAVDT2018,DL-Tracking-Survey} aims to accurately localize a model-free target, while robustly estimating a fitted bounding box on the target region. Given the wide variety of applications \cite{UAVApp1,UAVApp2}, vision-based methods for flying robots demand robust aerial visual trackers \cite{VisDrone2019-Chal,VisionMeetsDrones}. Generally speaking, aerial visual tracking can be categorized into videos captured from low-altitudes and medium/high-altitudes. Low-altitude aerial scenarios (e.g., UAV-123 \cite{UAV123}) look at medium or large objects in surveillance videos with limited viewing angles (similar to traditional tracking scenarios such as OTB \cite{OTB2013,OTB2015} or VOT \cite{VOT-2018,VOT-2019}). However, tracking a target in aerial videos captured from medium- (30$\sim$70 meters) and high-altitudes ($>$70 meters) has recently introduced extra challenges, including tiny objects, dense cluttered background, weather condition, wide aerial view, severe camera/object motion, drastic camera rotation, and significant viewpoint change \cite{VisDrone2018,VisDrone2019,UAVDT2018,UAVDT2019}. In most cases, it is arduous even for humans to track tiny objects in the presence of complex background as a consequence of limited pixels of objects. Fig.~\ref{fig:Scenarios} compares the two main categories of aerial visual tracking. Most objects in the first category (captured from low-altitude aerial views (10$\sim$30 meters)) are medium/large-sized and provide sufficient information for appearance modeling. The second one aims to track targets with few pixels involving complicated scenarios. \\
\indent Recent state-of-the-art trackers cannot provide satisfactory results for small object tracking since strategies to handle its challenges have not been considered. Besides, although various approaches have been proposed for small object detection \cite{SmallDetReview,ClusterNet,SOD-MTGAN}, there are limited methods to focus on aerial view tracking. These trackers \cite{UAV-Aberrance,UAV-Boundary,UAV-Distillation,UAV-KeyFilter,UAV-AutoTrack} are based on the \textit{discriminative correlation filters} (DCF) that have inherent limitations (e.g., boundary effect problem), and their performances are not competitive with modern trackers, despite extracting deep features. Besides, they cannot consider aspect ratio change despite being a critical characteristic for aerial view tracking. Therefore, the proposed method will narrow the gap between modern visual trackers with aerial ones. \\
\indent Tracking small objects involves major difficulties comprising lacking sufficient target information to distinguish it from background or distractors, much more possibility of locations (i.e., accurate localization requirement), and limited knowledge according to previous efforts. Motivated by the issues and also recent advances in small object detection, this paper proposes a \textit{Context-aware iOu-guided network for sMall objEct Tracking} (COMET). It exploits a multitask two-stream network to process target-relevant information at various scales and focuses on salient areas via attention modules. Given a rough estimation of target location by an online classification network \cite{ATOM}, the proposed network simultaneously predicts \textit{intersection-over-union} (IoU) and \textit{normalized center location error} (CLE) between the estimated \textit{bounding boxes} (BBs) and target. Moreover, an effective proposal generation strategy is proposed, which helps the network to learn contextual information. By using this strategy, the proposed network effectively exploits the representations of a target and its parts. It also leads to a better generalization of the proposed network to handle occlusion and viewpoint change for small object tracking from medium- and high-altitude aerial views. \\
\indent The contributions of the paper are summarized as the following two folds.\\
\indent \textbf{1) Offline Proposal Generation Strategy:} In offline training, the proposed method generates limited high-quality proposals from the reference frame. The proposed strategy provides context information and helps the network to learn target and its parts. Therefore, it successfully handles large occlusions and viewpoint changes in challenging aerial scenarios. Furthermore, it is just used in offline training to impose no extra computational complexity for online tracking.\\
\indent \textbf{2) Multitask Two-Stream Network:} COMET utilizes a multitask two-stream network to deal with challenges in small object tracking. First, the network fuses aggregated multi-scale spatial features with semantic ones to provide rich features. Second, it utilizes lightweight spatial and channel attention modules to focus on more relevant information for small object tracking. Third, the network optimizes a proposed multitask loss function to consider both accuracy and robustness. \\
\indent Extensive experimental analyses are performed to compare the proposed tracker with state-of-the-art methods on the well-known aerial view benchmarks, namely, UAVDT \cite{UAVDT2018}, VisDrone-2019 \cite{VisDrone2019}, Small-90 \cite{Small90Dataset}, and UAV-123 \cite{UAV123}. The results demonstrate the effectiveness of COMET for small object tracking purposes.\\
\indent The rest of the paper is organized as follows. In Section \ref{sec:2_RelatedWork}, an overview of related works is briefly outlined. In Section \ref{sec:3_PropMethod} and Section \ref{sec:4_ExpAnalyses}, our approach and empirical evaluation are presented. Finally, the conclusion is summarized in Section \ref{sec:5_Conc}.
\vspace{-3mm}
\section{Related Work} \label{sec:2_RelatedWork}
In this section, focusing on two-stream neural networks, modern visual trackers are briefly described.
Also, aerial visual trackers and some small object detection methods are summarized. 
\vspace{-3mm}
\subsection{Generic Object Tracking on Surveillance Videos} \label{sec:2.1}
Two-stream networks (a generalized form of \textit{Siamese neural networks} (SNNs)) for visual tracking were interested in \textit{generic object tracking using regression networks} (GOTURN) \cite{GOTURN}, which utilizes offline training of a network without any online fine-tuning during tracking. This idea continued by \textit{fully-convolutional Siamese networks} (SiamFC) \cite{SiamFC}, which defined the visual tracking as a general similarity learning problem to address limited labeled data issues. To exploit both the efficiency of the \textit{correlation filter} (CF) and CNN features, CFNet \cite{CFNet} provides a closed-form solution for end-to-end training of a CF layer. The work of \cite{Tripletloss} applies triplet loss on exemplar, positive instance, and negative instance to strengthen the feedback of back-propagation and provide powerful features. These methods could not achieve competitive performance compared with well-known DCF methods (e.g., \cite{CCOT,ECO}) since they are prone to drift problems; However, these methods provide beyond real-time speed.\\
\indent As the baseline of the well-known Siamese trackers (e.g., \cite{DaSiamRPN,SiamDW,CRPN,SiamMask,SiamRPN++}), the \textit{Siamese region proposal network} (SiamRPN) \cite{SiamRPN} formulates generic object tracking as local one-shot learning with bounding box refinement. \textit{Distractor-aware Siamese RPNs} (DaSiamRPN) \cite{DaSiamRPN} exploits semantic backgrounds, distractor suppression, and local-to-global search region to learn robust features and address occlusion and out-of-view. To design deeper and wider networks, the SiamDW \cite{SiamDW} has investigated various units and backbone networks to take full advantage of state-of-the-art network architectures. \textit{Siamese cascaded RPN} (CRPN) \cite{CRPN} consists of multiple RPNs that perform stage-by-stage classification and localization. SiamRPN++ method \cite{SiamRPN++} proposes a ResNet-driven Siamese tracker that not only exploits layer-wise and depth-wise aggregations but also uses a spatial-aware sampling strategy to train a deeper network successfully. SiamMask tracker \cite{SiamMask} benefits bounding box refinement and class agnostic binary segmentation to improve the estimated target region. \\
\indent Although the mentioned trackers provide both desirable performance and computational efficiency, they mostly do not consider background information and suffer from poor generalization due to lacking online training and update strategy. The ATOM tracker \cite{ATOM} performs classification and target estimation tasks with the aid of an online classifier and an offline IoU-predictor, respectively. First, it discriminates a target from its background, and then, an IoU-predictor refines the generated proposals around the estimated location. Similarly and based on a model prediction network, the DiMP tracker \cite{DiMP} learns a robust target model by employing a discriminative loss function and an iterative optimization strategy with a few steps. \\
\indent Despite considerable achievements on surveillance videos, the performance of modern trackers is dramatically decreased on videos captured from medium- and high-altitude aerial views; The main reason is lacking any strategies to deal with small object tracking challenges. For instance, the limited information of a tiny target, dense distribution of distractors, or significant viewpoint change leads to tracking failures of conventional trackers.
\vspace{-3mm}
\subsection{Detection/Tracking of Small Objects from Aerial View} \label{sec:2.2}
In this subsection, recent advances for small object detection and also aerial view trackers will be briefly described. \\
\indent Various approaches have been proposed to overcome shortcomings for small object detection \cite{SmallDetReview}. For instance, \textit{single shot multi-box detector} (SSD) \cite{SSD} uses low-level features for small object detection and high-level ones for larger objects. \textit{Deconvolutional single shot detector} (DSSD) \cite{DSSD} increases the resolution of feature maps using deconvolution layers to consider context information for small object detection. \textit{Multi-scale deconvolutional single shot detector for small objects} (MDSDD) \cite{MDSSD} utilizes several multi-scale deconvolution fusion modules to enhance the performance of small object detection. Also, \cite{SmallContextAttention} utilizes multi-scale feature concatenation and attention mechanisms to enhance small object detection using context information. SCRDet method \cite{SCRDet} introduces SF-Net and MDA-Net for feature fusion and highlighting object information using attention modules, respectively. Furthermore, other well-known detectors (e.g., YOLO-v3 \cite{YOLOv3}) exploit the same ideas, such as multi-scale feature pyramid networks, to alleviate their poor accuracy for small objects. \\
\indent On the other hand, developing specific methods for small object tracking from aerial view is still in progress, and there are limited algorithms for solving existing challenges. Current trackers are based on \textit{discriminative correlation filters} (DCFs), which provide satisfactory computational complexity and intrinsic limitations such as the inability to handle aspect ratio changes of targets. For instance, \textit{aberrance repressed correlation filter} \cite{UAV-Aberrance} (ARCF) proposes a cropping matrix and regularization terms to restrict the alteration rate of response map. To tackle boundary effects and improve tracking robustness,  \textit{boundary effect-aware visual tracker} (BEVT) \cite{UAV-Boundary} penalizes the object regarding its location, learns background information, and compares the scores of following response maps. Keyfilter-aware tracker \cite{UAV-KeyFilter} learns context information and avoids filter corruption by generating key-filters and enforcing a temporal restriction. To improve the quality of training set, \textit{time slot-based distillation algorithm} \cite{UAV-Distillation} (TSD) adaptively scores historical samples by a cooperative energy minimization function. It also accelerates this process by discarding low-score samples. Finally, the AutoTrack \cite{UAV-AutoTrack} aims to learn a spatio-temporal regularization term automatically. It exploits local-global response variation to focus on trustworthy target parts and determine its learning rate. The results of these trackers are not competitive to the state-of-the-art visual trackers (e.g., Siam-based trackers \cite{SiamRPN++,SiamMask}, ATOM \cite{ATOM}, and DiMP \cite{DiMP}). Therefore, the proposed method aims to narrow the gap between modern visual trackers and aerial view tracking methods, exploring small object detection advances.
\vspace{-3mm}
\section{Our Approach} \label{sec:3_PropMethod}
\vspace{-1mm}
\begin{figure}[!tb]
\centering
\includegraphics[width=12cm, height=5cm]{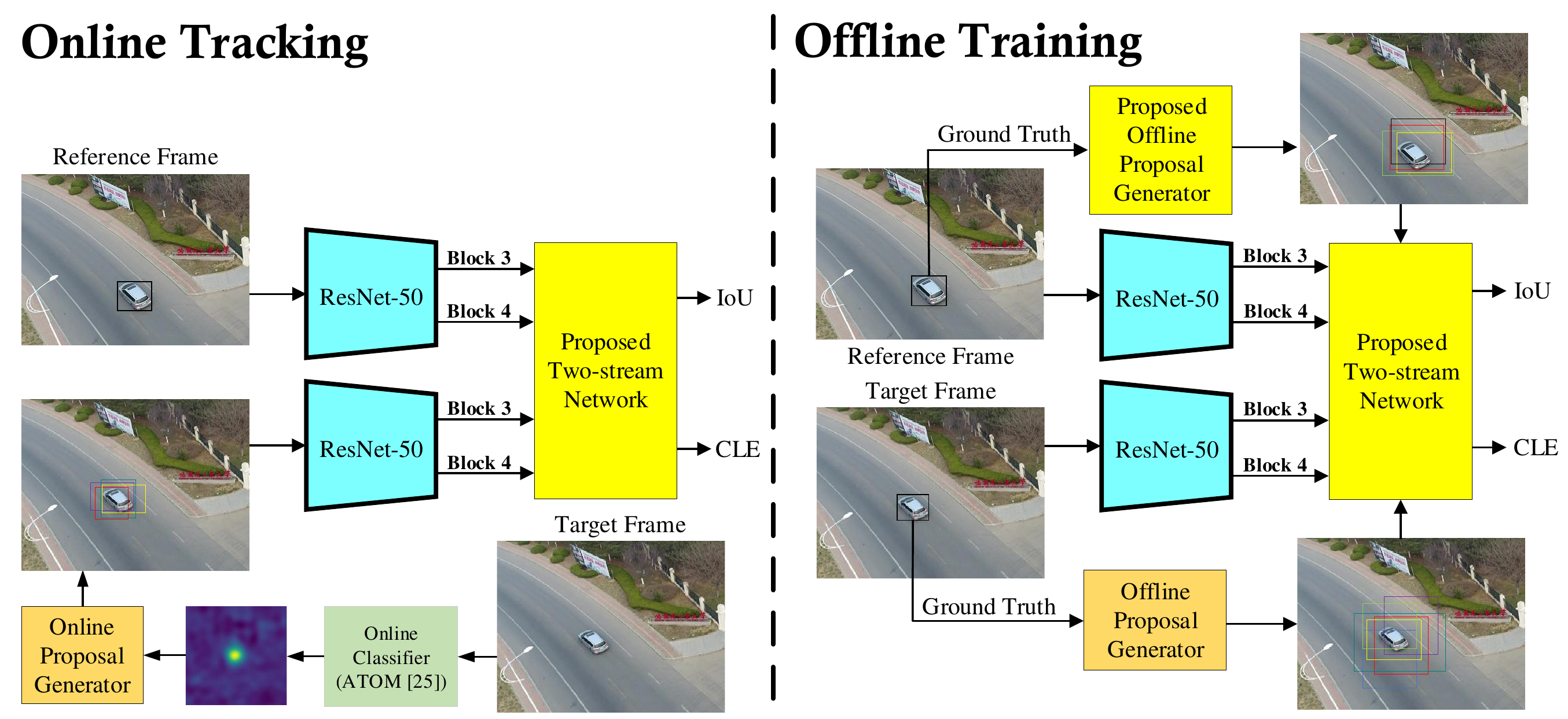}
\vspace{-2mm}
\caption{Overview of proposed method in offline training and online tracking phases.}
\label{fig:overview}
\vspace{-2mm}
\end{figure}
A key motivation of COMET is to solve the issues discussed in Sec. \ref{sec:1_Intro} and Sec. \ref{sec:2_RelatedWork} by adapting small object detection schemes into the network architecture for tracking purposes. The graphical abstract of proposed offline training and online tracking is shown in Fig.~\ref{fig:overview}. The proposed framework mainly consists of an offline proposal generation strategy and a two-stream multitask network, which consists of lightweight individual modules for small object tracking. Also, the proposed proposal generation strategy helps the network to learn a generalized target model, handle occlusion, and viewpoint change with the aid of context information. This strategy is just applied to offline training of the network to avoid extra computational burden in online tracking. This section presents an overview of the proposed method and a detailed description of the main contributions. 
\vspace{-4mm}
\subsection{Offline Proposal Generation Strategy} \label{sec:3.1}
\vspace{-3mm}
The eventual goal of proposal generation strategies is to provide a set of candidate detection regions, which are possible locations of objects. There are various category-dependent strategies for proposal generation \cite{RCNN,SSD,IoUNet}. For instance, the IoU-Net \cite{IoUNet} augments the ground-truth instead of using region proposal networks (RPNs) to provide better performance and robustness to the network. Also, the ATOM \cite{ATOM} uses a proposal generation strategy similar to \cite{IoUNet} with a modulation vector to integrate target-specific information into its network. \\
\indent Motivated by IoU-Net \cite{IoUNet} and ATOM \cite{ATOM}, an offline proposal generation strategy is proposed to extract context information of target from the reference frame. The ATOM tracker generates $N$ target proposals from the test frame ($\mathcal{P}_{t+\zeta}$), given the target location in that frame ($\mathcal{G}_{t+\zeta}$). Jittered ground-truth locations in offline training produce the target proposals. But, the estimated locations achieved by a simple two-layer classification network will be jittered in online tracking. The test proposals are generated according to ${IoU}_{Gt+\zeta}\triangleq IoU(\mathcal{G}_{t+\zeta},\mathcal{P}_{t+\zeta}) \geqslant \mathcal{T}_{1}$. Then, a network is trained to predict IoU values (${IoU}_{pred}$) between $\mathcal{P}_{t+\zeta}$ and object, given the BB of the target in the reference frame ($\mathcal{G}_{t}$). Finally, the designed network in the ATOM minimizes the mean square error of ${IoU}_{{G}_{t+\zeta}}$ and ${IoU}_{pred}$. \\
\indent In this work, the proposed strategy also provides target patches with background supporters from the reference frame (denoted as $\mathcal{P}_{t}$) to solve the challenging problems of small object tracking. Besides $\mathcal{G}_{t}$, the proposed method exploits $\mathcal{P}_{t}$ just in offline training. Using context features and target parts will assist the proposed network (Sec. \ref{sec:3.2}) in handling occlusion and viewpoint change problems for small objects. For simplicity, we will describe the proposed offline proposal generation strategy with the process of IoU-prediction. However, the proposed network predicts both IoU and center location error (CLE) of test proposals with target, simultaneously. \\
\indent An overview of the process of offline proposal generation for IoU-prediction is shown in Algorithm~\ref{alg:OfflineProposalStrategy}. The proposed strategy generates $(N/2)-1$ target proposals from the reference frame, which are generated as ${IoU}_{Gt}\triangleq IoU(\mathcal{G}_{t},\mathcal{P}_{t}) \geqslant \mathcal{T}_{2}$. Note that it considers $\mathcal{T}_{2} > \mathcal{T}_{1}$ to prevent drift toward visual distractors. The proposed tracker exploits this information (especially in challenging scenarios involving occlusion and viewpoint change) to avoid confusion during target tracking. The $\mathcal{P}_{t}$ and $\mathcal{G}_{t}$ are passed through the reference branch of the proposed network, simultaneously (Sec. \ref{sec:3.2}). In this work, an extended modulation vector has been introduced to provide the representations of the target and its parts into the network. That is a set of modulation vectors that each vector encoded the information of one reference proposal. To compute IoU-prediction, the features of the test patch should be modulated by the features of the target and its parts. It means that the IoU-prediction of $N$ test proposals is computed per each reference proposal. Thus, there will be $N^2/2$ IoU predictions. Instead of computing $N/2$ times of $N$ IoU-predictions, the extended modulation vector allows the computation of $N/2$ groups of $N$ IoU-predictions at once. Therefore, the network predicts $N/2$ groups of IoU-predictions by minimizing the mean square error of each group compared to ${IoU}_{Gt+\zeta}$. During online tracking, COMET does not generate $\mathcal{P}_{t}$ and just uses the $\mathcal{G}_{t}$ to predict one group of IoU-predictions for generated $\mathcal{P}_{t+\zeta}$. Therefore, the proposed strategy will not impose extra computational complexity in online tracking. 
\setlength{\textfloatsep}{4pt}
\begin{algorithm}[!t]
\algsetup{linenosize=\tiny}
\scriptsize
\caption{\textbf{:} Offline Proposal Generation}
\label{alg:OfflineProposalStrategy}
\textbf{Notations:} Bounding box $\mathcal{B}$ ($\mathcal{G}_{t+\zeta}$ for a test frame or $\mathcal{G}_{t}$ for a reference frame), IoU threshold $\mathcal{T}$ ($\mathcal{T}_{1}$ for a test frame or $\mathcal{T}_{2}$ for a reference frame), Number of proposals $\mathbb{N}$ ($N$ for a test frame or $(N/2)-1$ for a reference frame), Iteration number ($ii$), Maximum iteration ($max_{ii}$), A Gaussian distribution with zero-mean ($\mu = 0$) and randomly selected variance $\Sigma_r$ ($\mathcal{N}$), Bounding box proposals generated by a Gaussian jittering $\mathcal{P}$ ($\mathcal{P}_{t+\zeta}$ for a test frame or $\mathcal{P}_{t}$ for a reference frame) \\ 
\textbf{Input:} $\mathcal{B}$, $\mathcal{T}$, $\mathbb{N}$, $\Sigma_r$, $max_{ii}$   \\
\textbf{Output:} $\mathcal{P}$  \\
 \For{$i=1:\mathbb{N}$}{
 $ii = 0$,\\
 \Do{$(IoU(\mathcal{B},\mathcal{P}[i])<\mathcal{T})$  and  $(ii<max_{ii})$}{
  $\mathcal{P}[i]=\mathcal{B} + \mathcal{N}(\mu, \Sigma_r),$  \\
  $ii = ii + 1,$
 }
 }
\textbf{return} $\mathcal{P}$ 
\end{algorithm}
\vspace{-3mm}
\subsection{Multitask Two-Stream Network} \label{sec:3.2}
\begin{figure}[!tb]
\centering
\includegraphics[width=12cm, height=5.5cm]{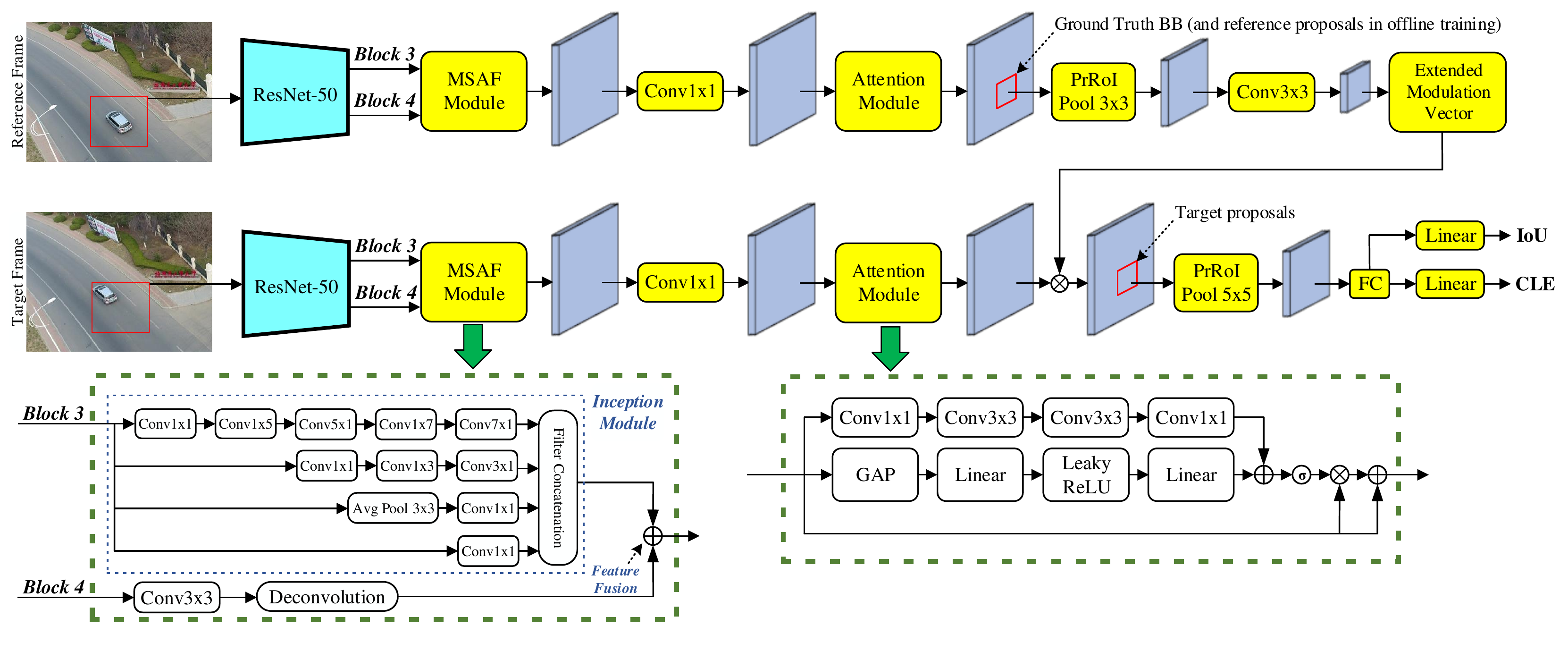}
\vspace{-5mm}
\caption{Overview of proposed two-stream network. MSAF denotes multi-scale aggregation and fusion module, which utilizes the InceptionV3 module in its top branch. For deconvolution block, a 3$\times$3 kernel with a stride of 2, input padding of 1, dilation value of 1, and output padding of 1 is used. After each convolution/fully-connected block, batch normalization and leaky ReLU are applied. Extended modulation vector allows COMET to learn targets and their parts in offline training. Also, the fully-connected block, global average pooling block, and linear layer are denoted as the FC, GAP, and linear, respectively.}
\label{fig:COMET_Net}
\vspace{6pt}
\end{figure}

Tracking small objects from aerial view involves extra difficulties such as clarity of target appearance, fast viewpoint change, or drastic rotations besides existing tracking challenges. This part aims to design an architecture that handles the problems of small object tracking by considering recent advances in small object detection. Inspired by \cite{IoUNet,ATOM,SCRDet,F-SSD,BAM}, a two-stream network is proposed (see Fig.~\ref{fig:COMET_Net}), which consists of multi-scale processing and aggregation of features, the fusion of hierarchical information, spatial attention module, and channel attention module. Also, the proposed network seeks to maximize the IoU between estimated BBs and the object while it minimizes their location distance. Hence, it exploits a multitask loss function, which is optimized to consider both the accuracy and robustness of the estimated BBs. In the following, the proposed architecture and the role of the main components are described. \\
\indent The proposed network has adopted ResNet-50 \cite{ResNet} to provide backbone features for reference and test branches. Following small object detection methods, features from block3 and block4 of ResNet-50 are just extracted to exploit both spatial and semantic features while controlling the number of parameters \cite{SmallDetReview,SCRDet}. Then, the proposed network employs a multi-scale aggregation and fusion module (MSAF). It processes spatial information via the InceptionV3 module \cite{InceptionV3} to perform factorized asymmetric convolutions on target regions. This low-cost multi-scale processing helps the network to approximate optimal filters that are proper for small object tracking. Also, semantic features are passed through the convolution and deconvolution layers to be refined and resized for feature fusion. The resulted hierarchical information is fused by an element-wise addition of the spatial and semantic feature maps. After feature fusion, the number of channels is reduced by 1$\times$1 convolution layers to limit the network parameters. Exploring multi-scale features helps the COMET for small objects that may contain less than $0.01$\% pixels of a frame. \\
\indent Next, the proposed network utilizes the bottleneck attention module (BAM) \cite{BAM}, which has a lightweight and simple architecture. It emphasizes target-related spatial and channel information and suppresses distractors and redundant information, which are common in aerial images \cite{SCRDet}. The BAM includes channel attention, spatial attention, and identity shortcut connection branches. In this work, the SENet \cite{SE-Res-PAMI} is employed as the channel attention branch, which uses global average pooling (GAP) and a multi-layer perceptron to find the optimal combination of channels. The spatial attention module utilizes dilated convolutions to increase the receptive field. It helps the network to consider context information for small object tracking. The spatial and channel attention modules answer to “where” the critical features are located and “what” relevant features are. Lastly, the identity shortcut connection helps for better gradient flow.\\
\indent After that, the proposed method generates proposals from the test frame. Also, it uses the proposed proposal generation strategy to extract the BBs from the target and its parts from the reference frame in offline training. These generated BBs are applied to the resulted feature maps and fed into a \textit{precise region of interest} (PrRoI) pooling layer \cite{IoUNet}, which is differentiable w.r.t. BB coordinates. The network uses a convolutional layer with a 3$\times$3 kernel to convert the PrRoI output to target appearance coefficients. Target coefficients are expanded and multiplied with the features of a test patch to merge the information of the target and its parts into the test branch. That is, applying target-specific information into the test branch by the extended modulation vector. Then, the test proposals ($\mathcal{P}_{t+\zeta}$) are applied to the features of the test branch and fed to a 5$\times$5 PrRoI pooling. Finally, the proposed network simultaneously predicts IoU and CLE of test proposals by optimizing a multitask loss function as $\mathcal{L}_{Net} = \mathcal{L}_{IoU} + \lambda\mathcal{L}_{CLE} \hspace{.1cm}$, where the $\mathcal{L}_{IoU}$, $\mathcal{L}_{CLE}$, and $\lambda$ represent the loss function for IoU-prediction head, loss function for the CLE-prediction head, and balancing hyper-parameter for loss functions, respectively. By denoting $i$-th IoU- and CLE-prediction values as ${IoU}^{(i)}$ and ${CLE}^{(i)}$, the loss functions are defined as
\vspace{-6pt}
\begin{gather}\label{Eq:L_IoU_CLE}
  \mathcal{L}_{IoU} = \frac{1}{N}\sum_{i=1}^{N}({IoU}_{G_{t+\zeta}}^{(i)}-{IoU}_{pred}^{(i)})^2 \hspace{.1cm}, \\
  \small{\mathcal{L}_{CLE} = \left\{\begin{matrix}
\frac{1}{N}\sum_{i = 1}^{N}\frac{1}{2}{({CLE}_{G_{t+\zeta}}^{(i)}-{CLE}_{pred}^{(i)})}^2 & |({CLE}_{G_{t+\zeta}}^{(i)}-{CLE}_{pred}^{(i)}|<1\\ 
\frac{1}{N}\sum_{i = 1}^{N}|({CLE}_{G_{t+\zeta}}^{(i)}-{CLE}_{pred}^{(i)})|- \frac{1}{2} & \hspace{.1cm} otherwise
\end{matrix}\right.},
\end{gather}
where the ${CLE}_{{G}_{t+\zeta}}=({\Delta x}_{{G}_{t+\zeta}}/{width}_{{G}_{t+\zeta}},{\Delta y}_{{G}_{t+\zeta}}/{height}_{{G}_{t+\zeta}})$ is the normalized distance between the center location of $\mathcal{P}_{t+\zeta}$ and $\mathcal{G}_{t+\zeta}$. For example, ${\Delta x}_{{G}_{t+\zeta}}$ is calculated as ${x}_{{G}_{t+\zeta}}-{x}_{{P}_{t+\zeta}}$. Also, the ${CLE}_{pred}$ (and ${IoU}_{pred}$) represents the predicted CLE (and the predicted IoU) between BB estimations ($\mathcal{G}_{t+\zeta}$) and target, given an initial BB in the reference frame. In offline training, the proposed network optimizes the loss function to learn how to predict the target BB from the pattern of proposals generation. \\
\begin{algorithm}[!b]
\algsetup{linenosize=\tiny}
\scriptsize
\caption{\textbf{:} Online Tracking}
\label{alg:OnlineTracking}
\textbf{Notations:} Input sequence ($\mathcal{S}$), Sequence length ($T$), Current frame ($t$), Rough estimation of bounding box ($\mathcal{B}_t^e$), Generated test proposals ($\mathcal{B}_t^p$), Concatenated bounding boxes ($\mathcal{B}_t^c$), Bounding box prediction ($\mathcal{B}_t^{pred}$), Step size ($\beta$), Number of refinements ($n$), Online classification network ($\mathbbm{Net}_{online}^{ATOM}$), Scale and center jittering ($Jitt$) with random factors, Network predictions ($IoU$ and $CLE$)\\
\textbf{Input:} $\mathcal{S}=\{I_0,I_1,...,I_T\}$, $\mathcal{B}_0=\{x_0,y_0,w_0,h_0\}$  \\
\textbf{Output:} $\mathcal{B}_t^{pred}$, $t\in\{1,...,T\}$     \\
 \For{$t=1:T$}{
  $\mathcal{B}_t^e = \mathbbm{Net}_{online}^{ATOM}(I_t)$ \\
  $\mathcal{B}_t^p = Jitt(\mathcal{B}_t^e)$ \\
  $\mathcal{B}_t^c = Concat(\mathcal{B}_t^e$, $\mathcal{B}_t^p$) \\
  \For{$i=1:n$}{
  $IoU$, $CLE$ = FeedForward($I_0$, $I_t$, $\mathcal{B}_0$, $\mathcal{B}_t^c$)   \\
  $\textbf{grad}_{\mathcal{B}_t^c}^{IoU}=[\frac{\partial IoU}{\partial x},\frac{\partial IoU}{\partial y},\frac{\partial IoU}{\partial w},\frac{\partial IoU}{\partial h}]$ \\
  $\mathcal{B}_t^c$ $\gets$ $\mathcal{B}_t^c+\beta\times[\frac{\partial IoU}{\partial x}.w,\frac{\partial IoU}{\partial y}.h,\frac{\partial IoU}{\partial w}.w,\frac{\partial IoU}{\partial h}.h]$, \\
  $\textbf{grad}_{\mathcal{B}_t^c}^{CLE}=[\frac{\partial CLE}{\partial x},\frac{\partial CLE}{\partial y},\frac{\partial CLE}{\partial w},\frac{\partial CLE}{\partial h}]$\\
  $\mathcal{B}_t^c$ $\gets$ $\mathcal{B}_t^c-\beta\times[\frac{\partial CLE}{\partial x}.w,\frac{\partial CLE}{\partial y}.h,\frac{\partial CLE}{\partial w},\frac{\partial CLE}{\partial h}]$
  }
  $\mathcal{B}_t^{K\times4}$ $\gets$ Select $K$ best $\mathcal{B}_t^c$ w.r.t. IoU-scores \\
  $\mathcal{B}_t^{pred} = Avg(\mathcal{B}_t^{K\times4})$
 }
\textbf{return} $\mathcal{B}_{t}^{pred}$
\end{algorithm}
\indent In online tracking, the target BB from the first frame (similar to \cite{SiamRPN,SiamMask,SiamRPN++,ATOM}) and also target proposals in the test frame passes through the network. As a result, there is just one group of CLE-prediction as well as IoU-prediction to avoid more computational complexity. In this phase, the aim is to maximize the IoU-prediction of test proposals using the gradient ascent algorithm and also to minimize its CLE-prediction using the gradient descent algorithm. Algorithm~\ref{alg:OnlineTracking} describes the process of online tracking in detail. This algorithm shows how the inputs are passed through the network, and BB coordinates are updated based on scaled back-propagated gradients. While the IoU-gradients are scaled up with BB sizes to optimize in a log-scaled domain (similar to \cite{IoUNet}), just $x$ and $y$ coordinates of test BBs are scaled up for CLE-gradients. It experimentally achieved better results compared to the scaling process for IoU-gradients. The intuitive reason is that the network has learned the normalized location differences between BB estimations and target BB. That is, the CLE-prediction is responsible for accurate localization, whereas the IoU-prediction determines the BB aspect ratio. After refining the test proposals ($N=10$ for online phase) for $n=5$ times, the proposed method selects the $K=3$ best BBs and uses the average of these predictions based on IoU-scores as the final target BB. \\
\vspace{-7mm}
\section{Empirical Evaluation} \label{sec:4_ExpAnalyses}
\vspace{-1mm}
In this section, the proposed method is evaluated on state-of-the-art benchmarks for small object tracking from aerial view: VisDrone-2019-test-dev (35 videos) \cite{VisDrone2019}, UAVDT (50 videos) \cite{UAVDT2018}, and Small-90 (90 videos) \cite{Small90Dataset}. Although the Small-90 dataset includes the challenging videos of the UAV-123 dataset with small objects, the experimental results on the UAV-123 \cite{UAV123} dataset (low-altitude UAV dataset (10$\sim$30 meters)) are also presented. Generally speaking, the UAV-123 dataset lacks varieties in small objects, camera motions, and real scenes \cite{UAVDT2019}. Moreover, traditional tracking datasets do not consist of challenges such as tiny objects, significant viewpoint changes, camera motion, and high density from aerial views. For instance, these datasets (e.g., OTB \cite{OTB2015}, VOT \cite{VOT-2018,VOT-2019}, etc.) mostly provide videos that captured by fixed or moving car-based cameras with limited viewing angles. For these reasons and our focus on tracking small objects on videos captured from medium- and high-altitudes, the proposed tracker (COMET) is evaluated on related benchmarks to demonstrated the motivation and major effectiveness of COMET for small object tracking.\\
\indent Experiments have been conducted three times, and the average results are reported. The proposed method is compared with state-of-the-art visual trackers in terms of precision, success, and \textit{normalized area-under-curve} (AUC) metrics \cite{OTB2015}. Note that all results have been produced by standard benchmarks with default thresholds (i.e., 20 pixels for precision scores, and 0.5 for the success scores) for fair comparisons. In the following, implementation details, ablation analyses, and state-of-the-art comparisons are presented.
\vspace{-3mm}
\subsection{Implementation Details}
The proposed method uses the ResNet-50 pre-trained on ImageNet \cite{ImageNet} to extract backbone features. For offline proposal generation, hyper-parameters are set to $N=16$ (test proposals number, $(N/2) = 8$ (seven reference proposal numbers plus reference ground-truth)), $\mathcal{T}_{1}=0.1$, $\mathcal{T}_{2}=0.8$, $\lambda=4$, and image sample pairs randomly selected from videos with a maximum gap of $50$ frames ($\zeta = 50$). From the reference image, a square patch centered at the target is extracted with the area of $5^2$ times the target region. Also, flipping and color jittering are used for data augmentation of the reference patch. To extract the search area, a patch (with the area of $5^2$ times the test target scale) with some perturbation in the position and scale is sampled from the test image. These cropped regions are then resized to the fixed size of 288$\times$288. The values for IoU and CLE are normalized to the range of $[-1, 1]$.\\
\indent The maximum iteration number $max_{ii}$ for proposal generation is $200$ for reference proposals and $20$ for test proposals. The weights of the backbone network are frozen, and other weights are initialized using \cite{He_init}. The training splits are extracted from the official training set (protocol II) of LaSOT \cite{LaSOT}, training set of GOT-10K \cite{GOT-10k}, NfS \cite{NfS}, and training set of VisDrone-2019 \cite{VisDrone2019} datasets. Note that the Small-112 dataset \cite{Small90Dataset} is a subset of the training set of the VisDrone-2019. Moreover, the validation splits of VisDrone-2019 and GOT-10K datasets have been used in the training phase. To train in an end-to-end fashion, the ADAM optimizer \cite{ADAM} is used with an initial learning rate of $10^{-4}$, weight decay of $10^{-5}$, and decay factor $0.2$ per $15$ epochs. The proposed network trained for $60$ epochs with a batch size of $64$ and $64000$ sampled videos per epoch. Also, the proposed tracker has been implemented using PyTorch, and the evaluations performed on an Nvidia Tesla V100 GPU with $16$ GB RAM. Finally, the parameters of the online classification network are set as the ATOM tracker \cite{ATOM}.
\begin{table}[b!]
\scriptsize
\caption{Ablation analysis of COMET regarding different components and feature fusion operations on UAVDT dataset.} 
\centering 
\begin{tabular}{P{2cm} P{2cm} P{1.5cm} P{1.5cm} P{1.5cm} P{1.5cm} P{1.5cm}} 
\hline \hline
Metric    & COMET & A1   & A2   & A3   & A4   & A5\\ \hline \hline
Precision & 88.7  & 87.2 & 85.2 & 83.6 & 88   & 85.3\\ 
Success   & 81    & 78   & 76.9 & 73.5 & 80.4 & 77.2\\ 
\hline
\end{tabular}
\label{Ablation}
\vspace{-.2cm}
\end{table}
\vspace{-7mm}
\subsection{Ablation Analysis}
A systematic ablation study on individual components of the proposed tracker has been conducted on the UAVDT dataset \cite{UAVDT2019} (see Table~\ref{Ablation}). It includes three different versions of the proposed network consisting of the networks without 1) “CLE-head”, 2) “CLE-head and reference proposals generation”, and 3) “CLE-head, reference proposals generation, and attention module”, referred to as A1, A2, and A3, respectively. Moreover, two other different feature fusion operations have been investigated, namely features multiplication (A4) and features concatenation (A5), compared to the element-wise addition of feature maps in the MSAF module (see Fig~\ref{fig:COMET_Net}).\\
\begin{figure}[b!]
\centering
\subfloat{\includegraphics[width=3cm,height=3cm]{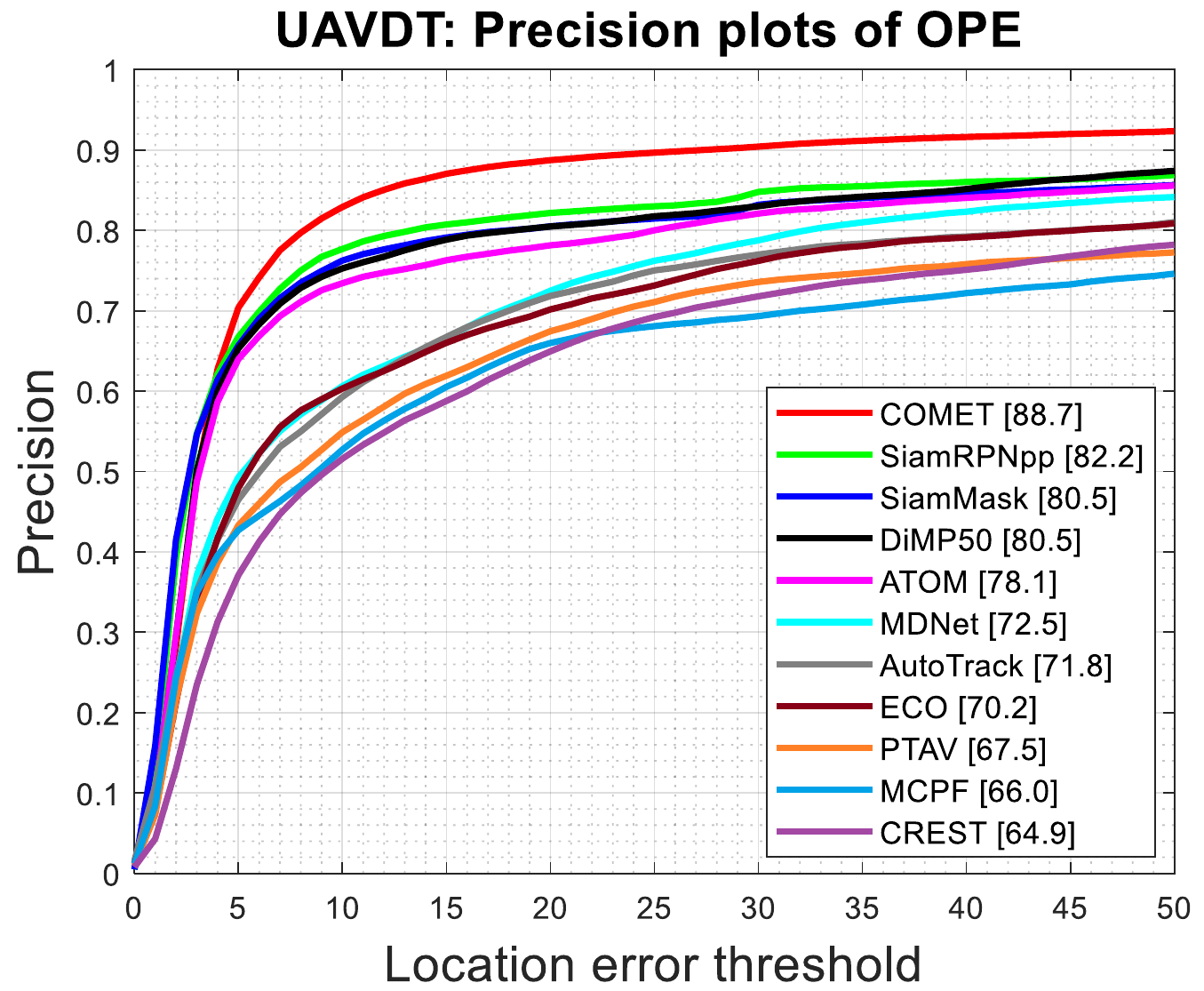}}
\hspace{0mm}
\subfloat{\includegraphics[width=3cm,height=3cm]{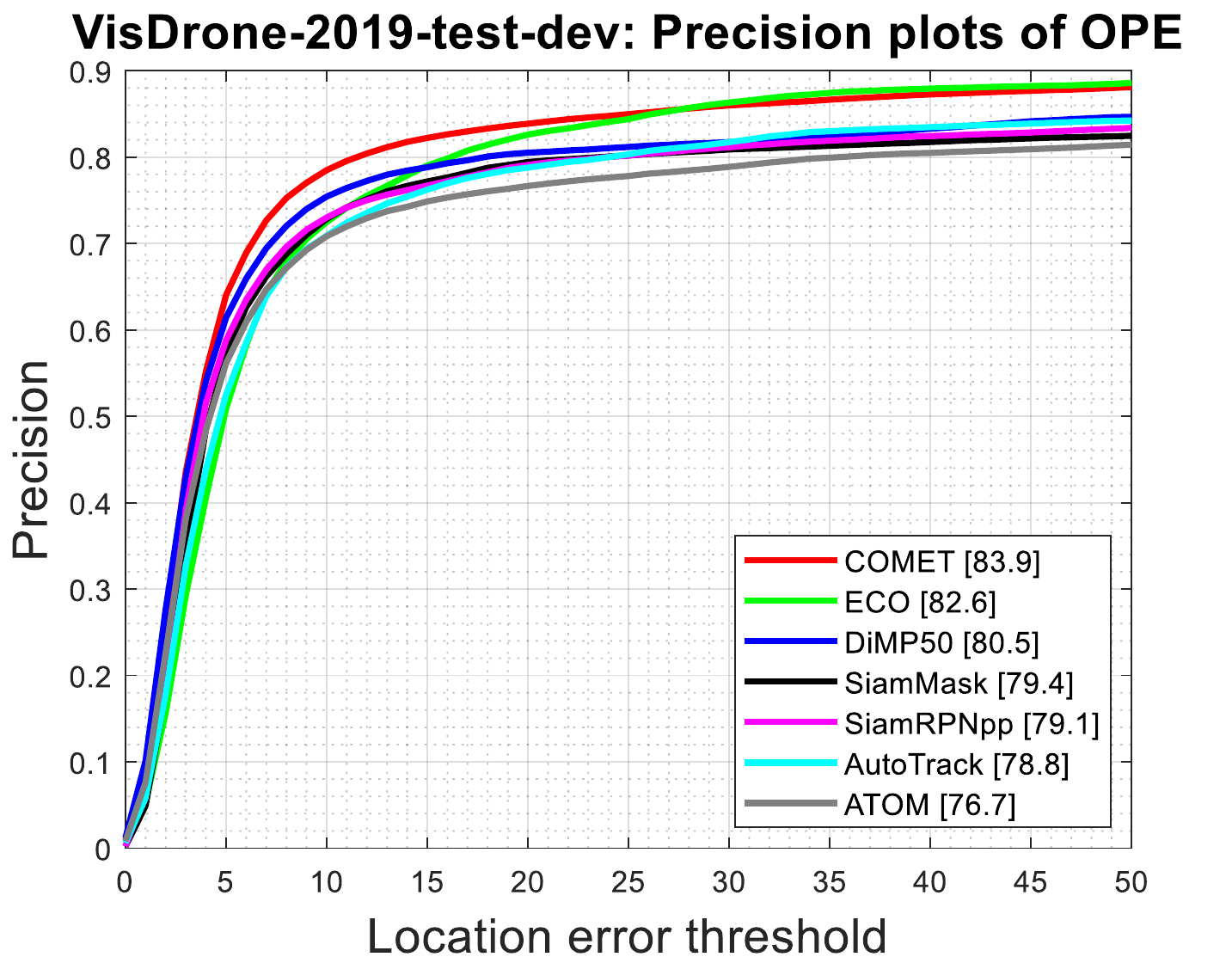}}
\hspace{-2mm}
\subfloat{\includegraphics[width=3cm,height=3.1cm]{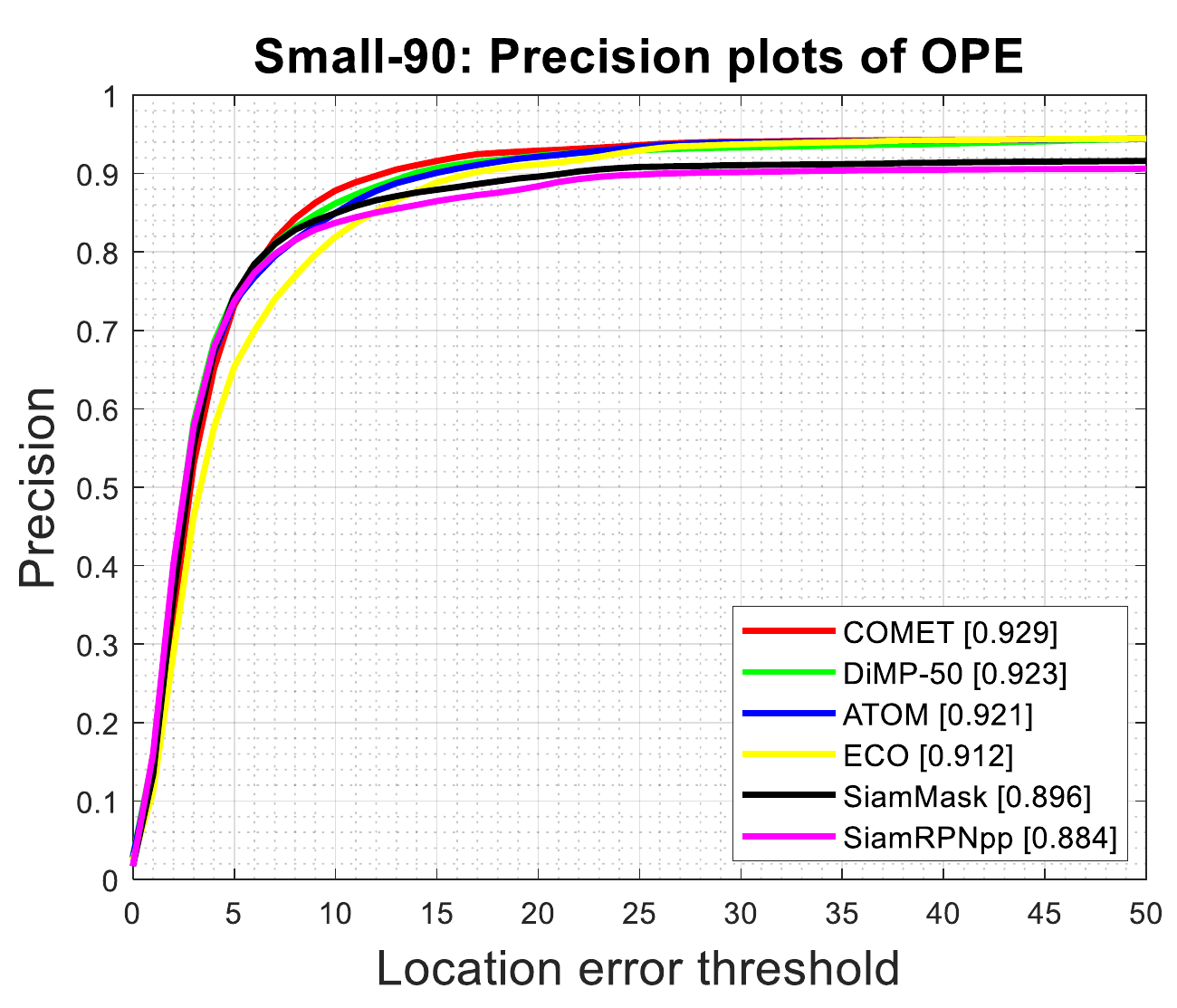}}
\hspace{0mm}
\subfloat{\includegraphics[width=3cm,height=3cm]{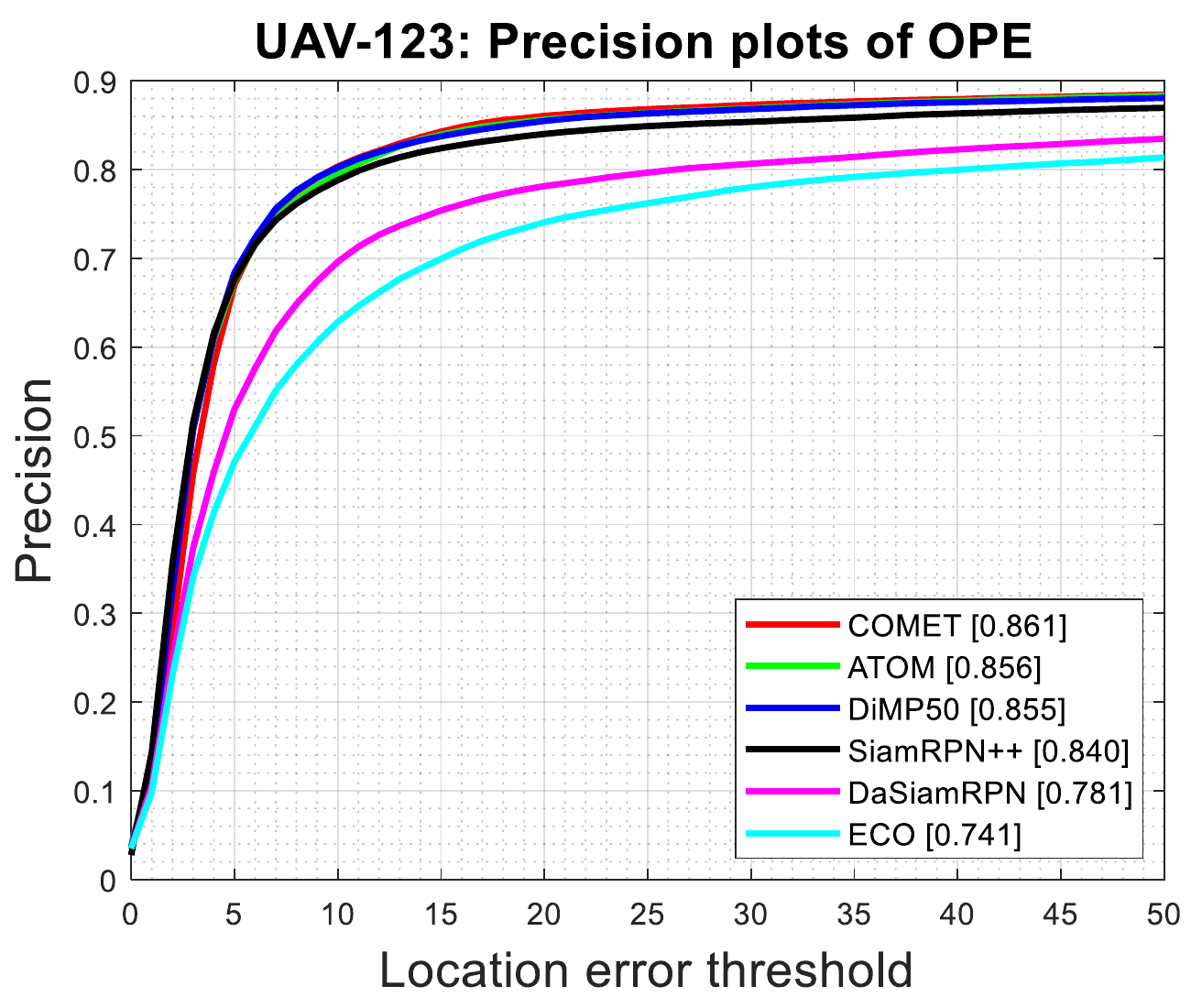}}
\vspace{-2mm}
\subfloat{\includegraphics[width=3cm,height=3cm]{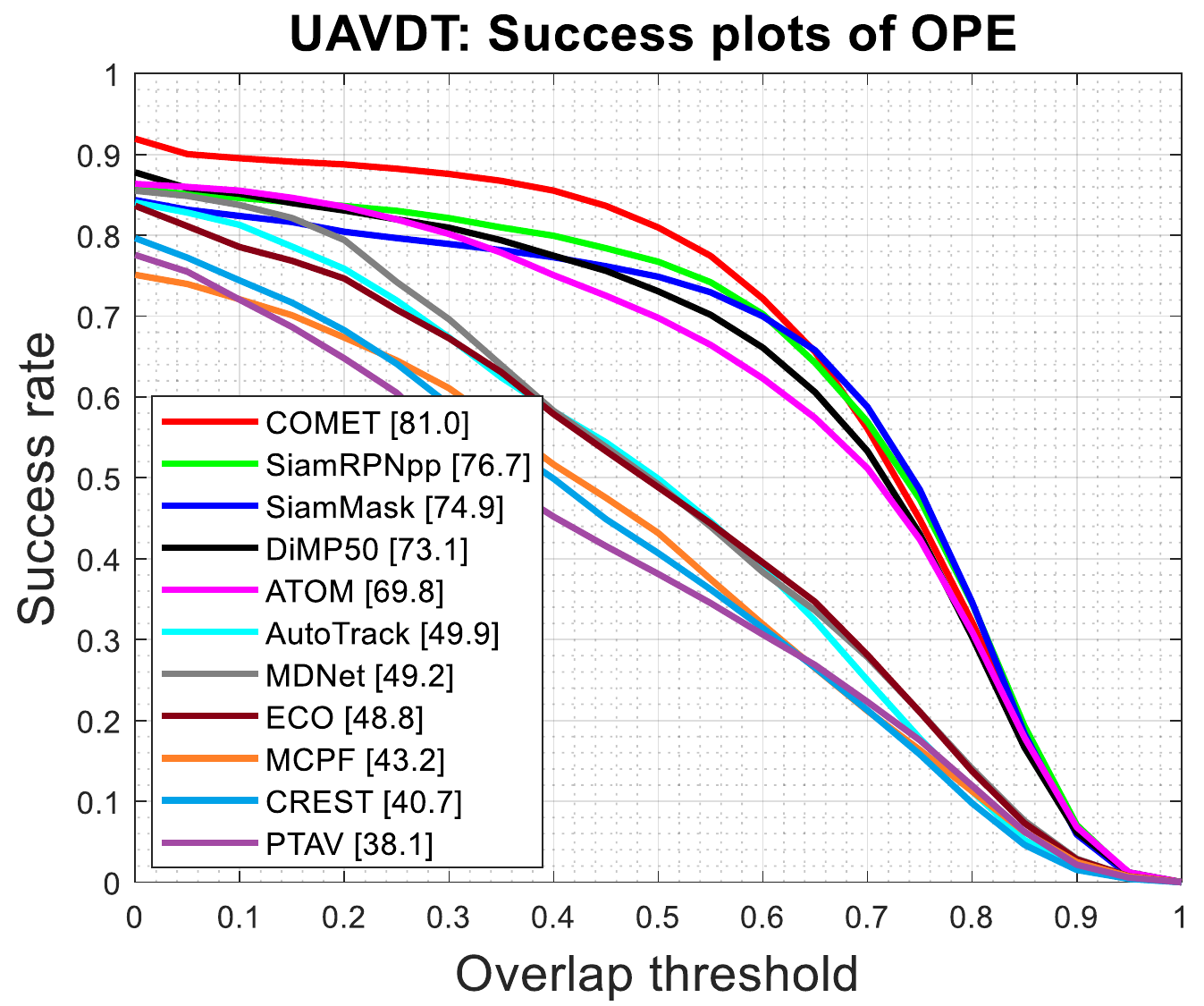}}
\hspace{0mm}
\subfloat{\includegraphics[width=3cm,height=3cm]{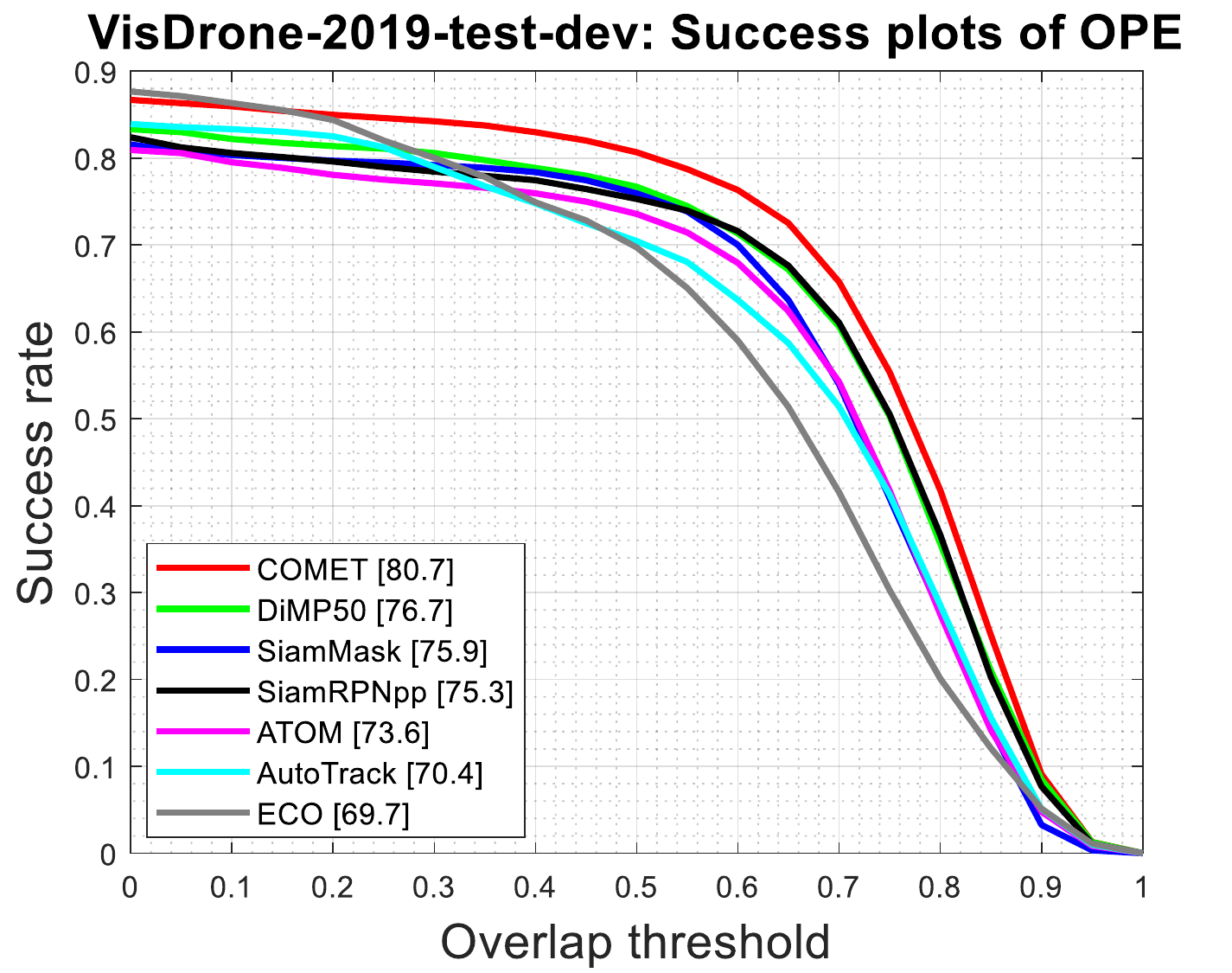}}%
\hspace{-2mm}
\subfloat{\includegraphics[width=3cm,height=3.05cm]{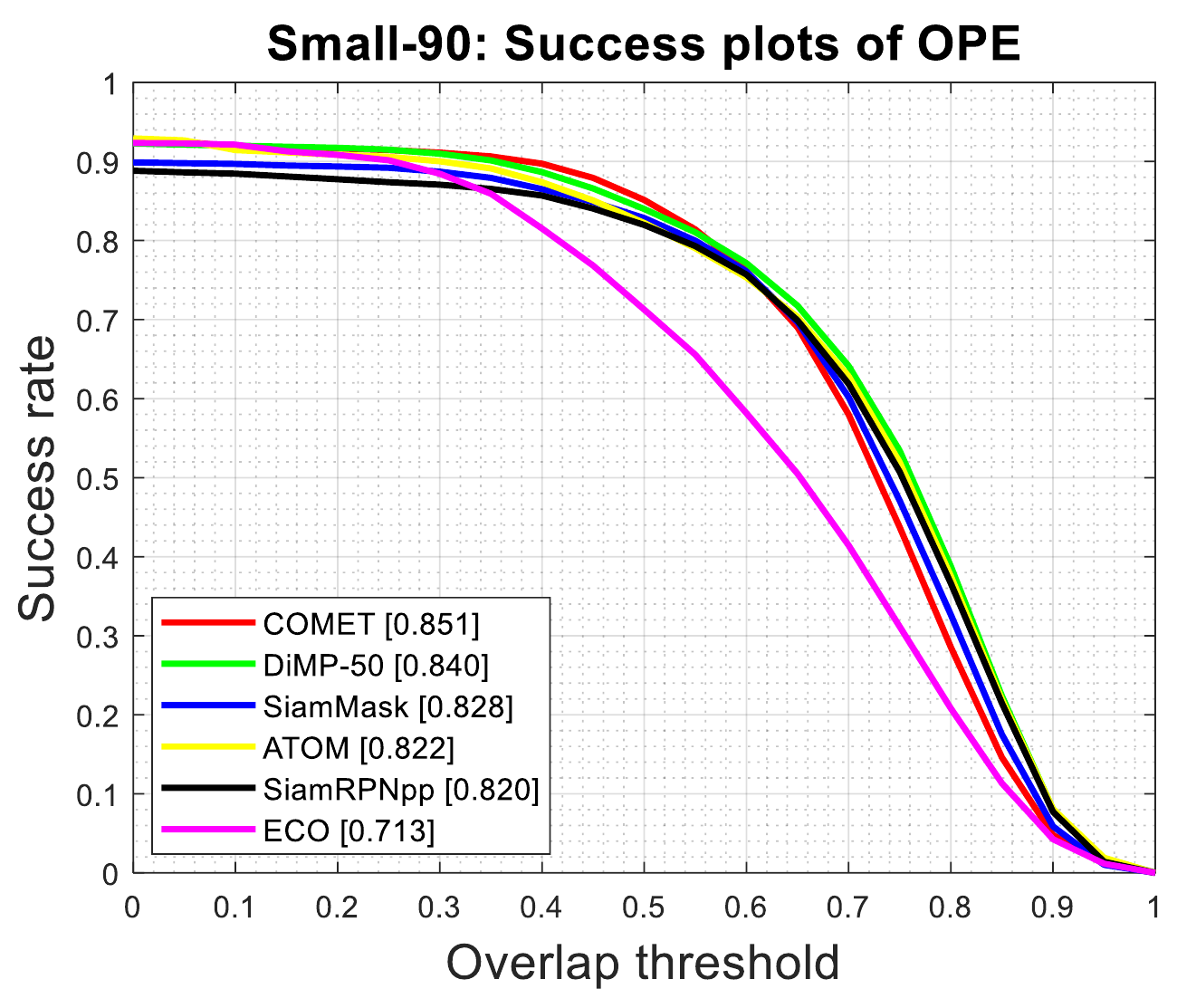}}
\hspace{0mm}
\subfloat{\includegraphics[width=3cm,height=3cm]{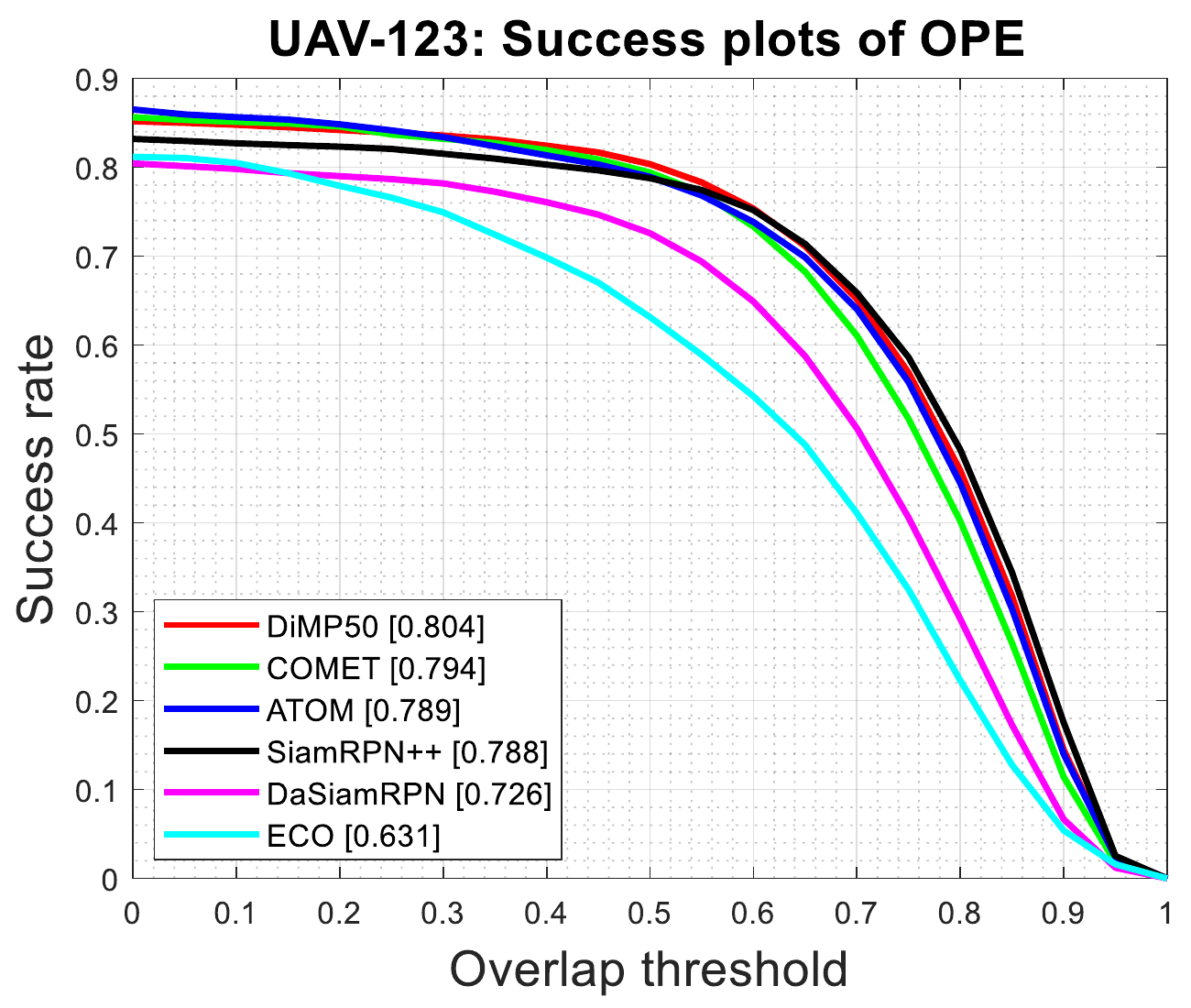}}
\vspace{-2mm}
\caption{Overall precision and success comparisons of the proposed method (COMET) with state-of-the-art tracking methods on UAVDT, VisDrone-2019-test-dev, Small-90, and UAV-123 datasets.}
\label{fig:OverallComp}
\vspace{-.5cm}
\end{figure}
\indent These experiments demonstrate the effectiveness of each component on tracking performance, while the proposed method has achieved $88.7\%$ and $81\%$ in terms of precision and success rates, respectively. According to these results, the attention module, reference proposal generation strategy, and CLE-head have improved the average of success and precision rates up to $2.5\%$, $1.55\%$, and $2.25\%$, respectively. Besides, comparing results of feature fusion operations demonstrate that the element-wise addition has provided the average of precision and success rates up to $0.65\%$ and $3.6\%$ compared to A4 and A5, respectively. Also, the benefit of feature addition previously has been proved in other methods such as \cite{ATOM}.
\vspace{-3mm}
\subsection{State-of-the-art Comparison}
\begin{figure}[!t]
\centering
\includegraphics[width=12cm, height=6cm]{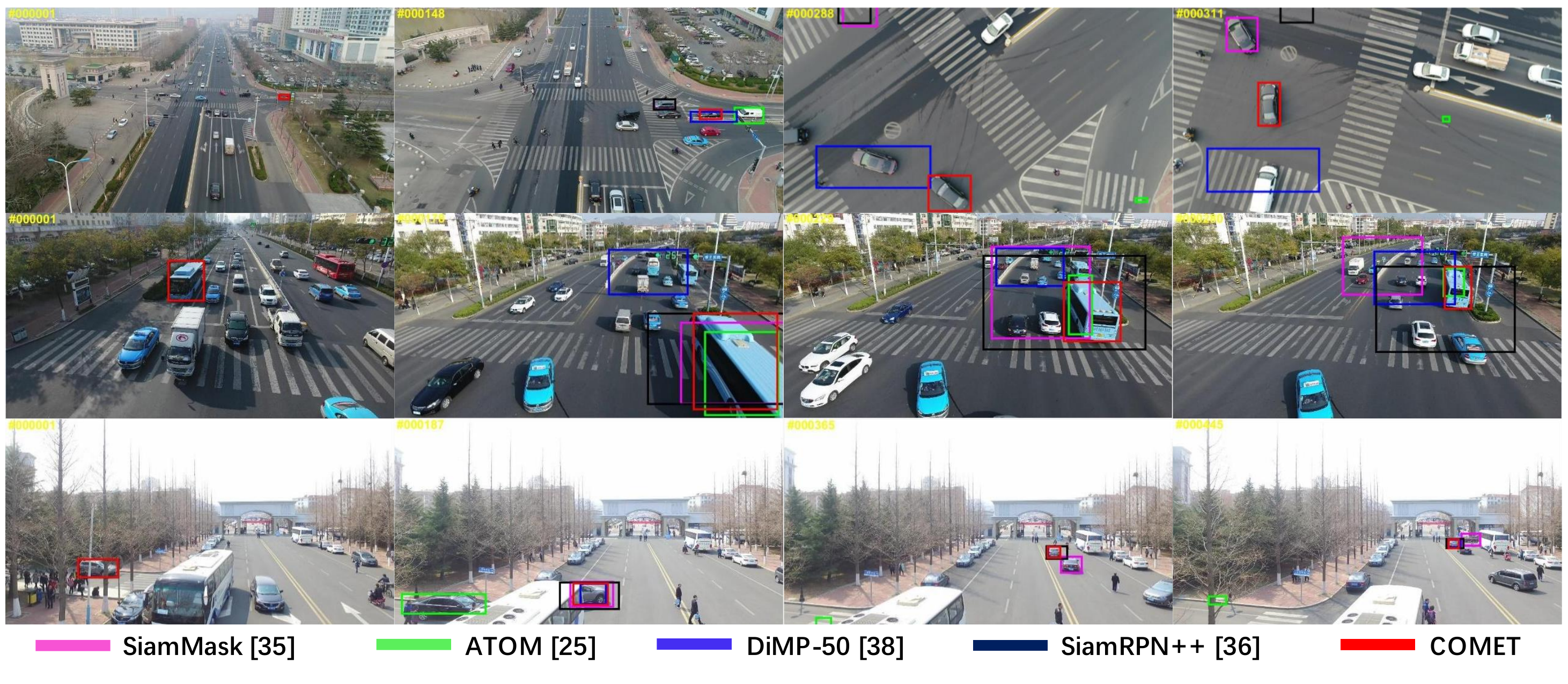}
\vspace{-3mm}
\caption{Qualitative comparison of proposed COMET tracker with state-of-the-art tracking methods on S1202, S0602, and S0801 video sequences from UAVDT dataset (top to bottom row, respectively).}
\label{fig:QualitativeComp}
\vspace{2mm}
\end{figure}

For quantitative comparison, the proposed method (COMET) is compared with state-of-the-art visual tracking methods including AutoTrack \cite{UAV-AutoTrack}, ATOM \cite{ATOM}, DiMP-50 \cite{DiMP}, SiamRPN$++$ \cite{SiamRPN++}, SiamMask \cite{SiamMask}, DaSiamRPN \cite{DaSiamRPN}, CREST \cite{CREST}, MDNet \cite{MDNet}, PTAV \cite{PTAV}, ECO \cite{ECO}, and MCPF \cite{MCPF} on aerial tracking datasets.\\ \indent Fig.~\ref{fig:OverallComp} shows the achieved results in terms of precision and success plots \cite{OTB2015}. According to these results, COMET outperforms top-performing visual trackers on three available challenging small object tracking datasets as well as the UAV-123 dataset. For instance, COMET has outperformed the SiamRPN++ and DiMP-50 trackers by $4.4\%$ and $3.2\%$ in terms of average precision metric, and $3.3\%$ and $3\%$ in terms of average success metric on all datasets, respectively. Compared to the baseline ATOM tracker, COMET has improved the average precision rate up to $10.6\%$, $7.2\%$ and $0.8\%$, while it increased the average success rate up to $11.2\%$, $7.1\%$ and $2.9\%$ on the UAVDT, VisDrone-2019-test-dev and Small-90 datasets, respectively. Although COMET slightly outperforms ATOM tracker on the UAV-123 (see Fig.~\ref{fig:Scenarios}), it achieved up to $6.2\%$ and $7\%$ improvements compared to it in terms of average precision and success metrics on small object tracking datasets. \\
\begin{table}[t!]
\scriptsize
\caption{Average speed of state-of-the-art trackers on UAVDT dataset.} 
\centering 
\begin{tabular}{P{2cm} P{1.6cm} P{1.4cm} P{2.1cm} P{1.6cm} P{1.8cm} P{1.4cm}} 
\hline \hline 
         & COMET & ATOM & SiamRPN++ & DiMP-50 & SiamMask & ECO \\ \hline \hline
Speed (FPS)    & 24    & 30   & 32        & 33      & 42       & 35 \\ 
\hline
\end{tabular}
\label{Speed}
\end{table}
\indent These results are mainly owed to the proposed proposal generation strategy and effective modules, which makes the network focus on relevant target (and its parts) information and context information. Furthermore, COMET runs at 24 \textit{frame-per-second} (FPS), while the average speed of other trackers on the referred machine is indicated in Table~\ref{Speed}. This satisfactory speed has been originated from considering different proposal generation strategies for offline \& online procedures and employing lightweight modules in the proposed architecture. \\
\begin{table}[b!]
\scriptsize
\caption{Attribute-based comparison of state-of-the-art visual tracking methods in terms of accuracy metric on the UAVDT dataset [\colorbox{green}{First} and \colorbox{yellow}{second} visual tracking methods are shown in color].} 
\centering 
\begin{tabular}{P{2.1cm} P{1cm} P{1cm} P{1cm} P{1cm} P{1cm} P{1cm} P{1cm} P{1cm} P{1cm}} 
\hline \hline 
Tracker & BC & CM & OM & SO & IV & OB & SV & LO & LT \\ \hline \hline
COMET & \cellcolor{green}83.8 & \cellcolor{green}86.1 & \cellcolor{green}90.6 & \cellcolor{green}90.9 & \cellcolor{yellow}88.5 & \cellcolor{yellow}87.7 & \cellcolor{green}90.2 & \cellcolor{green}79.6 & \cellcolor{yellow}96  \\ 
ATOM & 70.1 & 77.2 & 73.4 & 80.6 & 80.8 & 74.9 & 73 & 66 & 91.7  \\ 
AutoTrack & 61.1 & 66.5 & 63.1 & 80.9 & 76.8 & 73.2 & 63.6 & 52.1 & 87.8  \\ 
SiamRPN$++$ & \cellcolor{yellow}74.9 & 75.9 & \cellcolor{yellow}80.4 & 83.5 & \cellcolor{green}89.7 & \cellcolor{green}89.4 & \cellcolor{yellow}80.1 & 66.6 & 84.9  \\ 
SiamMask & 71.6 & 76.7 & 77.8 & \cellcolor{yellow}86.7 & 86.4 & 86 & 77.3 & 60.1 & 93.8  \\ 
DiMP-50 & 71.1 & \cellcolor{yellow}80.3 & 75.8 & 81.4 & 84.3 & 79 & 76.1 & \cellcolor{yellow}68.6 & \cellcolor{green}100  \\ 
ECO & 61.1 & 64.4 & 62.7 & 79.1 & 76.9 & 71 & 63.2 & 50.8 & 95.2  \\ 
MDNet & 63.6 & 69.6 & 66.8 & 78.4 & 76.4 & 72.4 & 68.5 & 54.7 & 93  \\ 
CREST & 56.2 & 62.1 & 55.8 & 74.2 & 69 & 65.6 & 56.7 & 49.7 & 71.2  \\ 
PTAV & 57.2 & 63.9 & 56.4 & 79.1 & 69.6 & 66.2 & 56.5 & 50.3 & 80.1  \\ 
MCPF & 51.2 & 59.2 & 55.3 & 74.5 & 73.1 & 73 & 55.1 & 42.5 & 74.1  \\ 
\hline
\end{tabular}
\label{AttComp}
\vspace{-.58cm}
\end{table}
\begin{table}[b!]
\scriptsize
\caption{Attribute-based comparison of state-of-the-art visual tracking methods in terms of AUC metric on the VisDrone-2019-test-dev dataset [\colorbox{green}{First} and \colorbox{yellow}{second} visual tracking methods are shown in color].} 
\centering 
\begin{tabular}{P{1.5cm} P{1cm} P{.75cm} P{.65cm} P{.65cm} P{.65cm} P{.75cm} P{.65cm} P{.65cm} P{.65cm} P{.75cm} P{.75cm} P{.65cm} P{.65cm}} 
\hline \hline 
Tracker & Overall & ARC & BC & CM & FM & FOC & IV & LR & OV & POC & SOB & SV & VC \\ \hline \hline
COMET & \cellcolor{green}64.5 & \cellcolor{green}64.2 & \cellcolor{green}43.4 & \cellcolor{green}62.6 & \cellcolor{green}64.9 & \cellcolor{green}56.7 & \cellcolor{green}65.5 & \cellcolor{yellow}41.8 & \cellcolor{green}75.9 & \cellcolor{green}62.1 & \cellcolor{green}42.8 & \cellcolor{green}65.8 & \cellcolor{green}70.4 \\ 
ATOM & 57.1 & 52.3 & 36.7 & 56.4 & 52.3 & 48.8 & 63.3 & 31.2 & 63 & 51.9 & 35.6 & 55.4 & 61.3  \\ SiamRPN$++$ & 59.9 & \cellcolor{yellow}58.9 & \cellcolor{yellow}41.2 & 58.7 & 61.8 & 55.1 & 63.5 & 36.4 & \cellcolor{yellow}69.3 & \cellcolor{yellow}58.8 & 39.6 & \cellcolor{yellow}59.9 & \cellcolor{yellow}67.8  \\ 
DiMP-50 & \cellcolor{yellow}60.8 & 54.5 & 40.6 & \cellcolor{yellow}60.6 & \cellcolor{yellow}62 & \cellcolor{yellow}55.8 & \cellcolor{yellow}63.6 & 32.7 & 62.4 & 56.8 & \cellcolor{yellow}39.8 & 59.7 & 66  \\ 
SiamMask & 58.1 & 57.8 & 38.5 & 57.2 & 60.8 & 49 & 56.6 & \cellcolor{green}46.5 & 67.5 & 52.9 & 37 & 59.4 & 65.1  \\
ECO & 55.9 & 56.5 & 38.3 & 54.2 & 52.1 & 46.8 & 59.9 & 36.6 & 61.5 & 51.6 & 38.6 & 52.7 & 62.4  \\ 
\hline
\end{tabular}
\label{AttComp2}
\end{table}
\indent The COMET has been evaluated according to various attributes of small object tracking scenarios to investigate its strengths and weaknesses. Table~\ref{AttComp} and Table~\ref{AttComp2} present the attribute-based comparison of visual trackers on the UAVDT and VisDrone datasets in terms of average precision and AUC metrics. Also, the overall AUC of trackers is shown in Table~\ref{AttComp2}. The comparisons are according to various attributes including: \textit{background clutter} (BC), \textit{illumination variation} (IV), \textit{scale variation} (SV), \textit{camera motion} (CM), \textit{object motion} (OM), \textit{small object} (SO), \textit{object blur} (OB), \textit{large occlusion} (LO), \textit{long-term tracking} (LT), \textit{aspect ratio change} (ARC), \textit{fast motion} (FM), \textit{partial occlusion} (POC), \textit{full occlusion} (FOC), \textit{low resolution} (LR), \textit{out-of-view} (OV), \textit{similar objects} (SOB), and \textit{viewpoint change} (VC). The results demonstrate the considerable performance of COMET compared to the state-of-the-art trackers. These tables also show that the COMET can successfully handle the occlusion and viewpoint change problems for small object tracking purposes. Compared to the DiMP-50, SiamRPN++, and SiamMask, COMET achieves improvements up to $9.5\%$, $7.4\%$, and $4.5\%$ for small object attribute, and $4.4\%$, $2.6\%$, and $5.3\%$ for viewpoint change attribute, respectively. Meanwhile, it has gained up to $5.7\%$, $5.9\%$, and $12.1\%$ improvement in average for occlusion attribute (i.e., average of FOC, POC, and LO) compare to the DiMP-50, SiamRPN++, and SiamMask, respectively. While the performance still can be improved based on IV, OB, LR, and LT attributes, COMET outperforms the ATOM by a margin up to $7.7\%$, $12.8\%$, $10.6\%$, and $4.3\%$ regarding these attributes, respectively. 
\indent The qualitative comparisons of visual trackers are shown in Fig.~\ref{fig:QualitativeComp}, in which the videos have been selected for more clarity. Further qualitative evaluations on various datasets and even YouTube videos are provided in the supplementary material. According to the first row of Fig.~\ref{fig:QualitativeComp}, COMET successfully models small objects on-the-fly considering complicated aerial view scenarios. Also, it provides promising results when the aspect ratio of target significantly changes. Examples of occurring out-of-view and occlusion are shown in the next rows of Fig.~\ref{fig:QualitativeComp}. By considering target parts and context information, COMET properly handles these problems existing potential distractors.  \\
\vspace{6mm}
\vspace{-1.1cm}
\section{Conclusion} \label{sec:5_Conc}
A context-aware IoU-guided tracker proposed that includes an offline reference proposal generation strategy and a two-stream multitask network. It aims to track small objects in videos captured from medium- and high-altitude aerial views. First, an introduced proposal generation strategy provides context information for the proposed network to learn the target and its parts. This strategy effectively helps the network to handle occlusion and viewpoint change in high-density videos with a broad view angle in which only some parts of the target are visible. Moreover, the proposed network exploits multi-scale feature aggregation and attention modules to learn multi-scale features and prevent visual distractors. Finally, the proposed multitask loss function accurately estimates the target region by maximizing IoU and minimizing CLE between the predicted box and object. Experimental results on four state-of-the-art aerial view tracking datasets and remarkable performance of the proposed tracker demonstrate the motivation and effectiveness of proposed components for small object tracking purposes.
\vspace{-.2cm}
\bibliographystyle{splncs}
\bibliography{egbib}

\end{document}